\documentclass[sigconf]{acmart}
\usepackage{algorithm}
\usepackage{algorithmic} 
\usepackage{multirow}
\usepackage{balance} 

\AtBeginDocument{%
  }

\copyrightyear{2025}
\acmYear{2025}
\setcopyright{acmlicensed}\acmConference[MM '25]{Proceedings of the 33rd ACM International Conference on Multimedia}{October 27--31, 2025}{Dublin, Ireland}
\acmBooktitle{Proceedings of the 33rd ACM International Conference on Multimedia (MM '25), October 27--31, 2025, Dublin, Ireland}
\acmDOI{10.1145/3746027.3755603}
\acmISBN{979-8-4007-2035-2/2025/10}

\begin{document}

\title{\textit{FreeInsert}: Personalized Object Insertion with Geometric and Style Control}

\author{Yuhong Zhang}
\affiliation{%
  \institution{ Shanghai Jiao Tong University}
  \city{Shanghai}
  \country{China}}
\email{rainbowow@sjtu.edu.cn}

\author{Han Wang}
\affiliation{%
  \institution{Shanghai Jiao Tong University}
  \city{Shanghai}
  \country{China}}
\email{esmuellert@sjtu.edu.cn}

\author{Yiwen Wang}
\affiliation{%
  \institution{Shanghai Jiao Tong University}
  \city{Shanghai}
  \country{China}}
\email{evonwang@sjtu.edu.cn}

\author{Rong Xie}
\affiliation{%
  \institution{Shanghai Jiao Tong University}
  \city{Shanghai}
  \country{China}}
\email{xierong@sjtu.edu.cn}

\author{Li Song}
\authornote{Corresponding author.}
\affiliation{%
  \institution{Shanghai Jiao Tong University}
  \city{Shanghai}
  \country{China}}
\email{song\_li@sjtu.edu.cn}

\renewcommand{\shortauthors}{Yuhong Zhang, Han Wang, Yiwen Wang, Rong Xie, and Li Song}

\begin{abstract}
  Text-to-image diffusion models have made significant progress in image generation, allowing for effortless customized generation. However, existing image editing methods still face certain limitations when dealing with personalized image composition tasks. First, there is the issue of lack of geometric control over the inserted objects. Current methods are confined to 2D space and typically rely on textual instructions, making it challenging to maintain precise geometric control over the objects. Second, there is the challenge of style consistency. Existing methods often overlook the style consistency between the inserted object and the background, resulting in a lack of realism. In addition, the challenge of inserting objects into images without extensive training remains significant. To address these issues, we propose \textit{FreeInsert}, a novel training-free framework that customizes object insertion into arbitrary scenes by leveraging 3D geometric information. Benefiting from the advances in existing 3D generation models, we first convert the 2D object into 3D, perform interactive editing at the 3D level, and then re-render it into a 2D image from a specified view. This process introduces geometric controls such as shape or view. The rendered image, serving as geometric control, is combined with style and content control achieved through diffusion adapters, ultimately producing geometrically controlled, style-consistent edited images via the diffusion model.
\end{abstract}

\ccsdesc[500]{Computing methodologies~Computer vision}

\keywords{Object Insertion, Image Generation, Diffusion Models, Personalized Generation}


\maketitle

\section{Introduction}
Recently, with the rise of diffusion models, significant progress has been made in image generation~\cite{dhariwal2021dbg,song2020ddim,rombach2022ldm,saharia2022photorealistic}. Notably, Stable Diffusion ~\cite{rombach2022ldm} and its variants have found widespread applications in personalized image synthesis~\cite{ruiz2023dreambooth,gal2022textualinversion,ye2023ipadapter}, image stylization~\cite{sohn2024styledrop,wang2023styleadapter,hertz2024stylealigned}, and image editing~\cite{hertz2022prompttoprompt,tumanyan2023plugandplay}. Consequently, employing diffusion models to address image-related tasks has become a mainstream trend.

In this paper, we investigate the task of personalized object insertion in image synthesis, which involves injecting a given object into an existing or newly generated background and merging the two representations to create a harmonious and visually appealing image. We aim for the shape and view of the inserted object to be controllable rather than randomly generated by the diffusion model. This task presents several challenges:
\begin{itemize}
    \item[a)] The basic attributes, such as the category and semantics of the object in the generated image should remain consistent with the object image.
    \item[b)] The geometric aspects, such as the view and pose of the object in the generated image, should meet user-specified requirements.
    \item[c)] The style of the object in the generated image should be consistent with the background to ensure realism.
    \item[d)] The non-target regions of the generated image should be consistent with the background and should not undergo any changes in color or content.
    \item[e)] A training-free method is expected to simplify the workflow and save computing resources.
\end{itemize}

Existing image composition methods often focus on addressing one or a few of these challenges. For instance, AnyDoor~\cite{chen2024anydoor} and Paint-by-Example~\cite{yang2023paintbyexample} insert images into background images by fine-tuning diffusion models, but these methods require substantial data for training and are more focused on generating realistic images, lacking in the ability to insert into artistic-style backgrounds. Methods like TF-ICON~\cite{lu2023tficon}, PrimeComposer~\cite{wang2024primecomposer}, and MagicInsert~\cite{ruiz2024magicinsert} aim to insert objects into given backgrounds. TF-ICON~\cite{lu2023tficon} and PrimeComposer~\cite{wang2024primecomposer} are training-free and operate by modifying attention mechanisms, while MagicInsert~\cite{ruiz2024magicinsert} uses LoRA~\cite{hu2021lora} fine-tuning for customized style transfer. However, these methods do not offer controllability over the pose and view of the object in the generated image, sometimes failing to meet user's requirements.

\begin{figure}[t]
\centering
\includegraphics[width=0.98\linewidth]{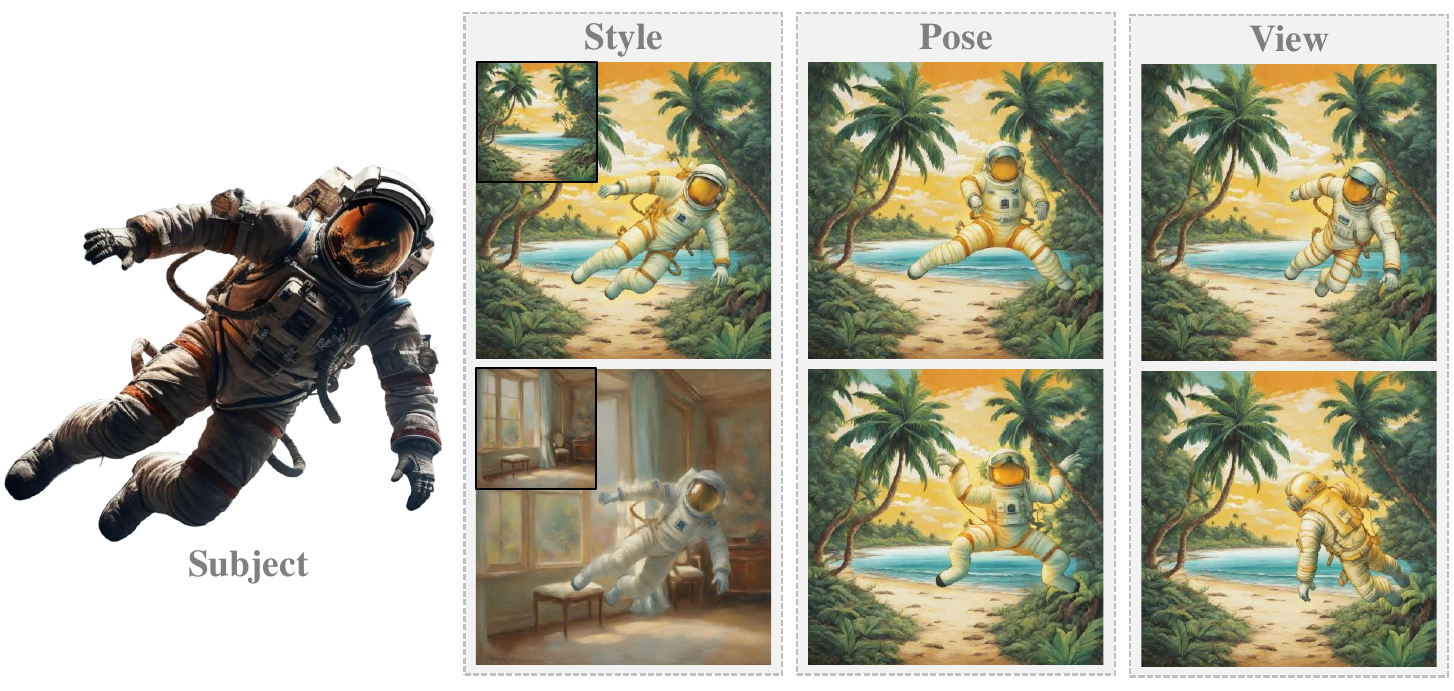} 
\vspace{-3mm}
\caption{Given an object image, our \textit{FreeInsert} can seamlessly integrate it into different background images with style awareness, while also enabling control over its pose and view.}
\label{fig:teaser}
\end{figure}

To address the aforementioned issues, we propose \textit{FreeInsert}, a training-free method for customized image composition. As shown in Figure \ref{fig:teaser}, given an object image, our method can insert the object into any scene while ensuring style consistency with the background and fidelity to the object. In addition, our method can customize the pose and view of objects in the synthetic image, thus achieving controllable object insertion. In order to achieve precise geometric control and ensure style consistency and object fidelity, we adopt a ``2D$\rightarrow$3D$\rightarrow$2D" solution. Specifically, we first lift the 2D image to 3D space and perform fine geometric editing in 3D space, such as pose. Then, we render it back to 2D domain according to the user's required view. Finally, we use the object image, background and rendered image as conditions for image synthesis. Through our designed geometric control from the rendered image, content control from the object image and style control from the background, our method can finally achieve geometrically controllable and style-consistent object insertion.

Our main contributions are summarized as follows:
\begin{itemize}
    \item We introduce a new task of personalized object insertion, where the object is inserted into the background image with controllable geometry and consistent style.
    \item We propose \textit{FreeInsert}, a method to address the personalized object insertion task, enabling controllable style, pose, and view.
    \item Our method avoids the need for training or fine-tuning, leveraging existing models to solve the task of personalized object insertion. 
    \item Qualitative and quantitative results demonstrate the superiority of our method in the personalized object insertion task.
\end{itemize}

\section{Related Work}
\subsection{Image Composition}
Image composition task, where objects are inserted into existing scenes, was initially explored using GAN~\cite{zhang2019shadowgan}. With the development of diffusion models, current methods have started to achieve this through fine-tuning or training additional adapters. Paint-By-Example~\cite{ye2023ipadapter} uses Stable Diffusion (SD)~\cite{rombach2022ldm} and classifier-free guidance to generate reference images within the masked regions of the target image, but it only retains semantic information and offers low fidelity to the original object's identity. AnyDoor~\cite{chen2024anydoor} leverages DINO-V2~\cite{oquab2023dinov2} to extract identity features and uses high-frequency details from the reference image to capture image details. ObjectDrop~\cite{winter2024objectdrop} employs a dataset of ``counterfactual" image pairs, showing scenes before and after object removal, to train SD for object deletion or insertion. These methods often rely on extensive data for training and are less effective in handling style adaptation issues. TF-ICON~\cite{lu2023tficon} is a training-free cross-domain image synthesis method. It uses a special inversion technique to invert images and injects object information into the background image through composite attention. PrimeComposer~\cite{wang2024primecomposer} controls attention weights at different noise levels to preserve the object's appearance while naturally blending it with the background. MagicInsert~\cite{ruiz2024magicinsert} introduces a style-aware method for inserting objects into target images, which has significant advantages in style consistency. However, the aforementioned methods often lack geometric control over the generated objects, sometimes failing to meet user's requirements.

\subsection{Image Stylization}
Image stylization is a classic task in the field of image processing. Early work utilized convolutional neural networks or GANs to enhance the generative capabilities of models~\cite{gatys2016stylecnn,huang2017adain,park2019styleattention}. With the development of diffusion models, image stylization has started to leverage these models. StyleDrop~\cite{sohn2024styledrop} controls the style of generated images by learning style identifiers through two-stage fine-tuning. StyleAdapter~\cite{wang2023styleadapter} trains an adapter for stylization. These methods all rely on additional fine-tuning or training. Stylealigned~\cite{hertz2024stylealigned} introduces style control from an attention perspective and uses AdaIN~\cite{huang2017adain} to normalize attention queries and key values. However, this process is prone to content leakage from the reference style. InstantStyle~\cite{wang2024instantstyle} achieves image style transfer without fine-tuning by injecting reference style features into specific cross-attention layers of the IP-Adapter~\cite{ye2023ipadapter}. We also adopt a diffusion adapter to control the style of the target region in the generated image.

Additionally, Image Sculpturing~\cite{yenphraphai2024imagesculpture} is  similar to our work. We both follow the ``2D$\rightarrow$3D$\rightarrow$2D" solution.  However, Image Sculpturing  needs additional  training to learn the attribute of the objects and  does not consider style consistency.  What's more, it is a solution for the object editing in the same image, while we focus on object insertion across the different images.

\begin{figure*}[t]
\centering
\includegraphics[width=0.80\textwidth]{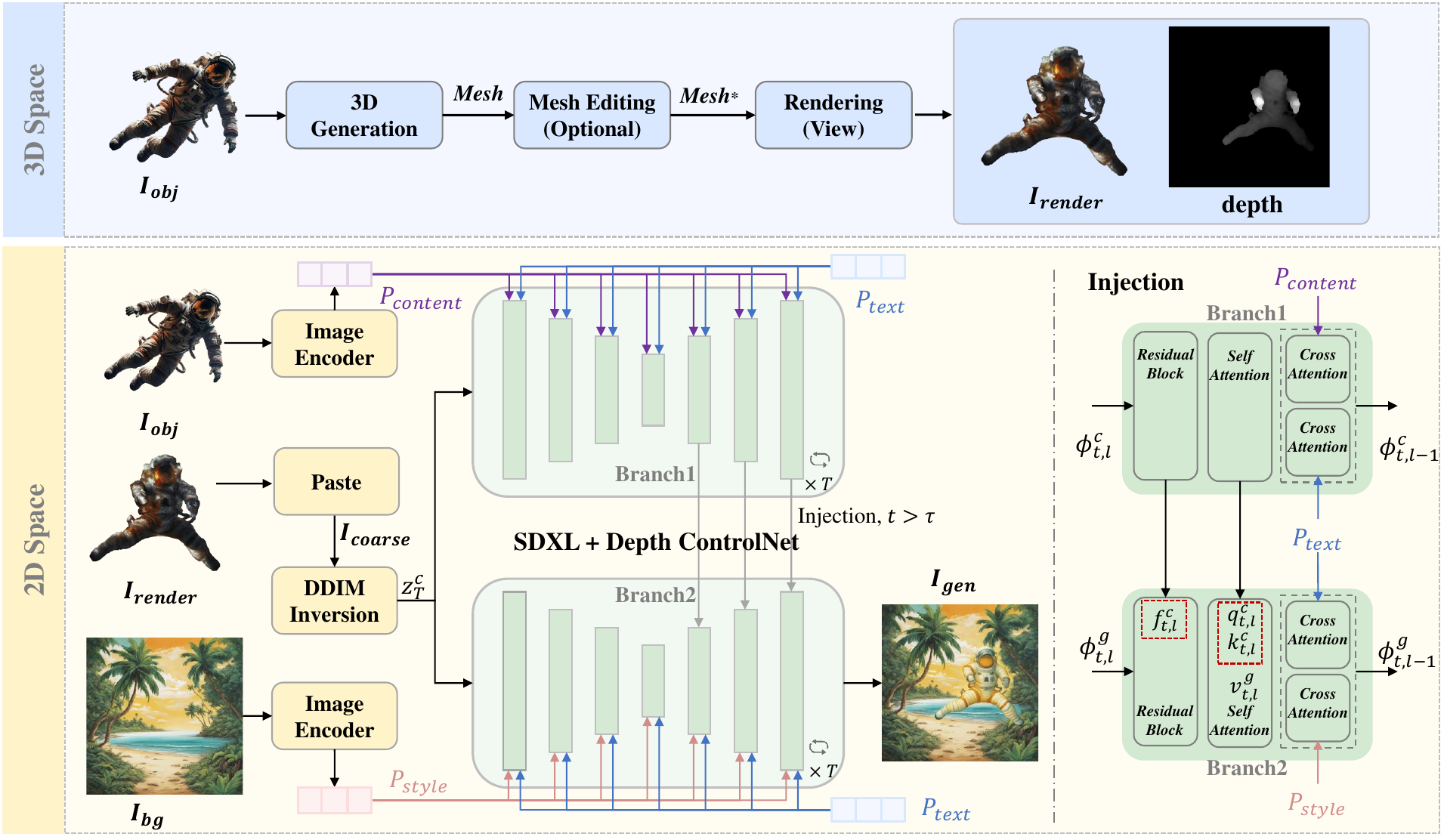} 
\vspace{-3mm}
\caption{Pipeline overview. The pipeline of \textit{FreeInsert} can be primarily divided into two parts: 3D Edit and 2D Generation. Given an object image, we first use the Image-to-3D model to obtain a 3D mesh, then edit the mesh with precise control and select the corresponding view to render to obtain the geometric conditions. In 2D space, we use the geometric conditions ($I_{render}$ and depth), the background image ($I_{bg}$) and the object image ($I_{obj}$) to control the diffusion model for desired image generation. Particularly, we first paste $I_{render}$ onto $I_{bg}$ to get $I_{coarse}$, which contains the geometry we expect, but not the texture details and style. So we use it as the geometric condition. Branch1 is the $I_{coarse}$ denoising process and Branch2 is our generation process. We inject Branch1's feature into Branch2 to make geometric control. In addition, the depth controlnet part and noise blending are omitted and the depth controlnet is consistent with ControlNet~\cite{zhang2023controlnet}.}
\label{fig:overview}
\end{figure*}

\subsection{Image-to-3D}
Recently, 3D generative models have achieved unprecedented advancements. DreamFusion~\cite{poole2022dreamfusion} pioneered the introduction of Score Distillation Sampling (SDS), utilizing 2D diffusion models to create 3D assets from text prompts or images. Zero-123~\cite{liu2023zero123} leverages extensive 3D data, training a diffusion model for novel view synthesis by altering the camera view of a given single RGB image. Magic123~\cite{qian2023magic123} proposes a 3D generation model optimized with 2D stable diffusion guidance and 3D Zero-123 guidance. DreamGaussian~\cite{tang2023dreamgaussian} is the first to introduce Gaussian Splatting into the single-view 3D reconstruction domain. TripoSR~\cite{tochilkin2024triposr}, based on the LRM~\cite{hong2023lrm} architecture, proposed a new method for fast feedforward 3D generation from a single image. In our method, we use the Image-to-3D model provided in threestudio~\cite{liu2023threestudio} with Zero-123 guidance and SDS optimization. More high-quality 3D generative models can also be integrated into our framework.

\section{Method}
\subsection{Preliminaries}
\textbf{Stable Diffusion.}
Stable Diffusion comprises a Variational Autoencoder (VAE) and a diffusion model. Given an input image $I$, the VAE encoder projects the image into the latent space, obtaining the latent code $z_0$. The diffusion process occurs in the latent space. This diffusion process consists of a forward pass and a reverse pass. In the forward pass, noise is continuously added to $z_0$ in T steps, i.e., $z_t = \sqrt{\alpha_t} z_0 + \sqrt{1-\alpha_t} \epsilon$.
The reverse pass employs a network $\epsilon_\theta$ to predict the noise added at any step $t$ for denoising. Finally, the VAE decoder reconstructs the denoised latent back into the pixel space. The optimization objective of the diffusion process is:
\begin{equation}
\label{lossdiffusion}
\mathcal{L} = \mathcal{E}_{z_0,t,P,\epsilon \sim \mathcal{N}}\left[||\epsilon-\epsilon_{\theta}(z_t, t, P)||_2^2\right]
\end{equation}
where $P$ is the additional condition, such as text prompt. In our work, we employ the Stable Diffusion-XL (SDXL)~\cite{podell2023sdxl} as the pre-trained text-to-image diffusion model. Building upon the Latent Diffusion Model (LDM)~\cite{rombach2022ldm}, SDXL incorporates a cascaded refinement model to significantly enhance the quality of the generated images.

\subsection{Overview}
In this section, we delineate the primary task addressed in this work, the key challenges encountered, and the pipeline of our proposed \textit{FreeInsert}.

\subsubsection{Task Setting}
Given an object image $I_{obj}$ and a background image $I_{bg}$, along with a user-specified position to insert and desired view or geometry, our objective is to insert the object into the background while the view and pose of the object in the generated result $I_{gen}$ should meet the user's requirements. This task faces many challenges: 1) the essential attributes such as category and semantics of the object in $I_{gen}$ should align with those of the object image $I_{obj}$; 2) the geometric aspects such as the view and pose of the object in $I_{gen}$ must meet the user's specifications; 3) the style of the object in $I_{gen}$ should be consistent with the $I_{bg}$ to ensure natural; 4) the non-target regions in $I_{gen}$ should remain consistent with $I_{bg}$ and not undergo any alterations.

\subsubsection{Pipeline Overview}
To address the aforementioned challenges, we propose \textit{FreeInsert}, a training-free framework designed for seamless object insertion into arbitrary background images with controllable geometry and style perception. As illustrated in Figure \ref{fig:overview}, our framework can be primarily divided into two parts: 3D Editing and 2D Generation. To achieve precise geometric and view control of objects in \( I_{gen} \), we adopt an intuitive approach by first lifting the object into 3D space and performing physically feasible edits. Specifically, given an object image \( I_{obj} \), we employ a 3D generation model~\cite{liu2023threestudio} to create a 3D asset. Users can then manipulate the mesh in 3D space to meet their specific requirements. Subsequently, the 3D mesh is rendered from the desired viewpoint, producing a rendered image \( I_{render} \). 

However, due to limitations in current 3D generative models, the texture details in \( I_{render} \) are often coarse and contain artifacts. Directly compositing \( I_{render} \) onto the background typically results in undesirable artifacts and stylistic inconsistencies between \( I_{gen} \) and \( I_{bg} \). To address this, we leverage a diffusion model to generate a high-quality image under the geometric guidance of \( I_{render} \), ensuring the preservation of geometric structure while maintaining stylistic consistency with \( I_{bg} \). Concretely, we first composite \( I_{render} \) onto \( I_{bg} \) to obtain \( I_{coarse} \), and then apply DDIM inversion to derive the noise \( z_T^c \). This noise \( z_T^c \) is fed into the diffusion model, which collaborates with geometry, content, and style control to produce the final desired result.

\subsection{Editing in 3D Space}
To apply precise geometric control to the object image and select the expected view, we first lift the object into 3D space using a 3D generation model~\cite{liu2023threestudio}. In the 3D space, we edit the generated mesh to achieve the desired pose or shape as specified by the user. This step can be skipped if no changes to the object's pose or shape are needed. After obtaining the edited mesh, we select the desired view for rendering, which results in an image that incorporates geometric control. The overall process is illustrated in Figure \ref{fig:overview}. However, since the rendered image $I_{render}$ is often rough and contains many artifacts, directly pasting it onto the background yields suboptimal results and fails to meet user expectations. Additionally, if the object image and background are not in the same style domain, style inconsistency issues may arise. Therefore, we use the rendered image solely as a geometric control condition to guide the diffusion model in generating higher-quality images.

\subsection{Controllable Generation in 2D Space}
In the 2D generation stage, we aim to control the generative model based on geometry, content, and style to ensure that the geometric shape or view aligns with the rendered image $I_{render}$, the generated image in the target regions matches the object image $I_{obj}$, and the style is consistent with the background image $I_{bg}$. We first paste $I_{render}$ onto $I_{bg}$ to generate $I_{coarse}$. $I_{coarse}$ is send to DDIM inversion to get inverted noise $z_T^c$. $z_T^c$ is passed through a diffusion model with geometry, content and style control to generate edited images. The controllable generation algorithm is shown in Algorithm \ref{alg:algorithm}. 

\subsubsection{Geometry Control}
To preserve geometric structure in $I_{gen}$, we first utilize a depth-based ControlNet~\cite{zhang2023controlnet} to enforce geometric shapes. Unlike general depth control, we can render depth maps directly from 3D models without requiring separate depth estimation. Since existing 3D generation models often produce coarse geometric shapes, depth control from the rendered depth provides only overall control, lacking more detailed control. It is claimed that the feature maps and self-attention maps in T2I diffusion models carry semantic and geometric information in \cite{tumanyan2023plugandplay}. This inspires us to make geometric control by manipulating the spatial features and self-attention features within the diffusion model. As illustrated in Figure \ref{fig:overview}, we first use DDIM inversion~\cite{song2020ddim} to get the inverted noise from $I_{coarse}$. Then we employ feature injection to better maintain geometric shapes. Specifically, we extract feature maps from residual blocks and self-attention maps from transformer blocks in the $I_{coarse}$ denoising process (Branch1) and inject them into the denoising process of generation (Branch2). In other words, at denoising step $t > \tau$, we override the spatial features $\{f_{t,l}^g\}$ and self-attention features $\{q_{t,l}^g, k_{t,l}^g\}$ of the diffusion model in the generation branch (Branch2) with $\{f_{t,l}^c\}$ and $\{q_{t,l}^c, k_{t,l}^c\}$ from $I_{coarse}$ denoising branch (Branch1), thus achieving geometric information injection. Here, $\tau$ is the injection threshold. 

\begin{algorithm}[tb]
\caption{Controllable Generation}
\label{alg:algorithm}
\textbf{Input}: Object image $I_{obj}$, rendered image $I_{render}$, background image $I_{bg}$, prompt $P_{text}$, depth map $D$, mask $M$, thresholds $\tau_f$, $\tau_q$, $\tau_k$ , diffusion model with depth controlnet $\epsilon_{\theta}$\\
\textbf{Output}: Composition result $I_{gen}$
\begin{algorithmic}[1] 
\STATE $I_{coarse} =\text{paste}(I_{render},I_{bg})$
\STATE $z_0^{c} =\text{Encoder}(I_{coarse})$, $z_0^{bg} = \text{Encoder}(I_{bg})$
\STATE ${z}_0^c, \hdots {z}_T^c \leftarrow \text{DDIM-Inversion}(I_{coarse}, P_{text}, D)$
\STATE $z_T^g \leftarrow z_T^c$ 
\FOR{$t=T$ to $1$}
\STATE $f_t^c$, $q_t^c$, $k_t^c \leftarrow \epsilon_{\theta}(z_t^c, P_{text}, I_{obj}, D, t)$ \hfill $\triangleright$ Branch1\\
\STATE $f_t^g$, $q_t^g$, $k_t^g \leftarrow \epsilon_{\theta}(z_t^g,P_{text},I_{bg},D,t)$ \hfill $\triangleright$ Branch2\\
\STATE \textbf{if} {$t > \tau_f$} \textbf{then} $f_t^g \leftarrow f_t^c$ \textbf{end if}
\STATE \textbf{if} {$t > \tau_q$} \textbf{then} $q_t^g \leftarrow q_t^c$ \textbf{end if}
\STATE \textbf{if} {$t > \tau_k$} \textbf{then} $k_t^g \leftarrow k_t^c$ \textbf{end if}  \hfill $\triangleright$ feature injection\\
\STATE $\epsilon_{t-1}^g \leftarrow \epsilon_{\theta}(z_t^g,P,D, t;f_t^g,q_t^g,q_t^g)$
\STATE $z_{t-1}^g \leftarrow \text{DDIM-Sampling}(z_t^g, \epsilon_{t-1}^g)$
\STATE $z_{t-1}^g = M \odot z_{t-1}^g+ (1-M) \odot z_{t-1}^{bg}$, $z_{t-1}^{bg} = \sqrt{\alpha_{t-1}} z_0^{bg} + \sqrt{1-\alpha_{t-1}} \epsilon$  \hfill $\triangleright$ noise blending\\
\ENDFOR
\STATE $I_{gen} \leftarrow \text{Decoder}(z_0^g)$
\end{algorithmic}
\end{algorithm}

\begin{figure*}[]
\centering
\includegraphics[width=0.88\textwidth]{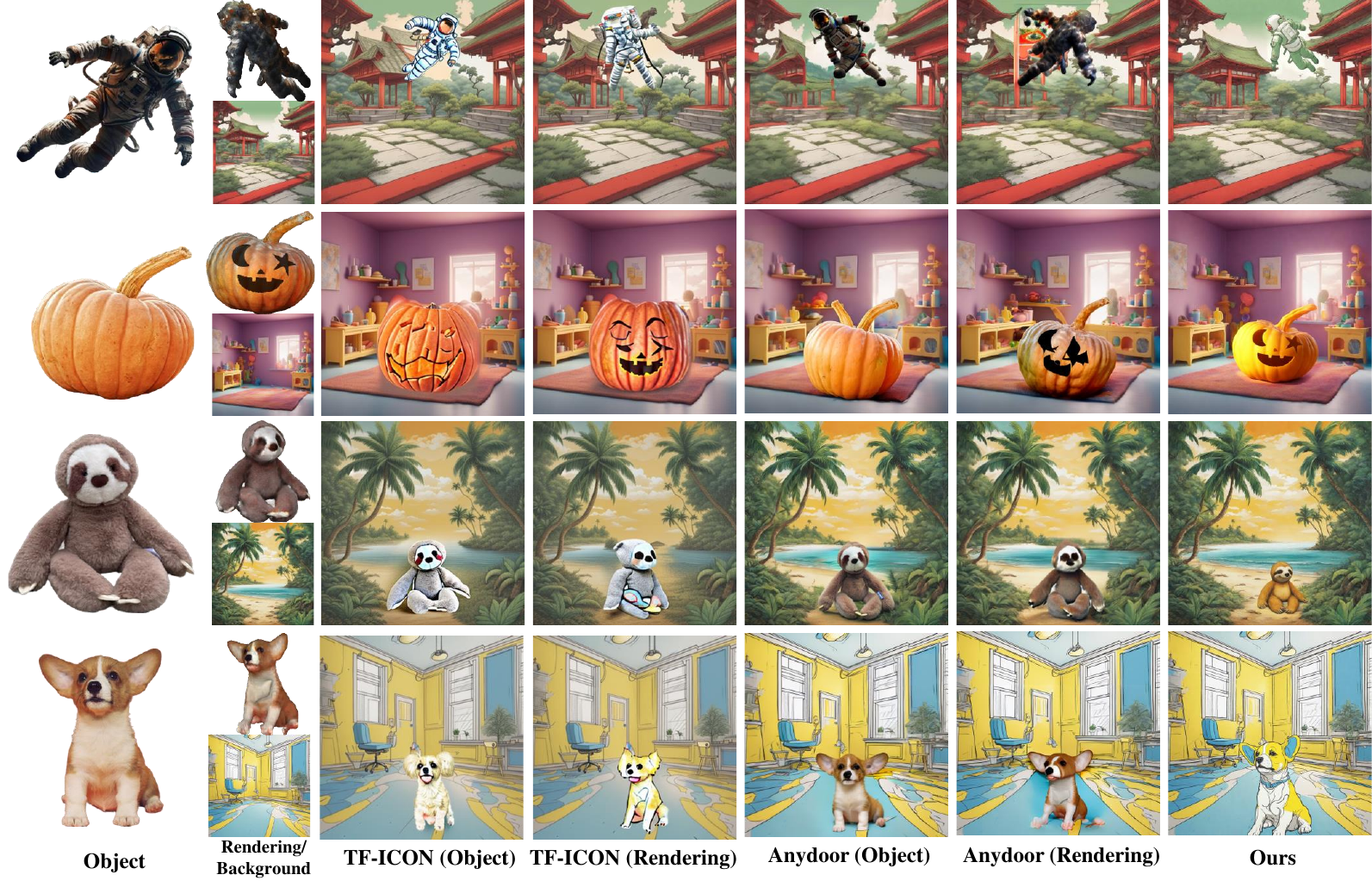} 
\vspace{-3mm}
\caption{Visual comparisons. Our \textit{FreeInsert} has significant advantages over TF-ICON~\cite{lu2023tficon} and AnyDoor~\cite{chen2024anydoor} in terms of background preservation, style consistency, and geometric controllability for the personalized image composition task.}
\label{fig:compare}
\end{figure*}

\subsubsection{Content Control}
For customized generation tasks, a common approach is to use DreamBooth~\cite{ruiz2023dreambooth} to finetune the model with an object-specific identifier. This method requires additional training for each object and typically needs 3-5 images to prevent overfitting. However, we aim to explore a training-free method to achieve controllable object insertion. Thus, we combine fine-grained descriptive prompts with an adapter-based control mechanism to achieve this. To obtain detailed descriptions of objects, we use the vision language model LLaVA~\cite{liu2024llava} to generate prompts $P_{text}$. Then we employ IP-Adapter~\cite{ye2023ipadapter} to inject the original object embedding $P_{content}$ into the diffusion model through cross-attention. Given that diffusion models focus on content generation in the early stages and style generation in the later stages, we inject $P_{content}$ only into Branch1. Through feature injection, Branch1 can effectively propagate content information to Branch2 for generation as shown in Figure \ref{fig:overview}.

\subsubsection{Style Control}
To maintain style consistency between the generated object and the background, we use IP-Adapter~\cite{ye2023ipadapter} to extract style embedding $P_{style}$ from $I_{bg}$ and then injected $P_{style}$ into the generation process through cross-attention. This process can enable style control of the generated image. Additionally, to seamlessly merge the edited object into the background image while ensuring that other elements in the background remain unchanged, we employ the noise blending strategy. Specifically, during each denoising step $t$, we preserve the background areas's original state with mask. The masked (edited) regions are then mixed with the unmasked (original) background to retain the unedited background:
\begin{equation}
    z_t = M \odot z_t + (1-M) \odot z_t^{bg}
\end{equation}
where, $z_t^{bg}$ is obatianed by adding noise to $z_0^{bg}$,$z_0^{bg}$ is obtained by VAE encoder with $I_{bg}$ input.

\begin{figure*}[]
\centering
\includegraphics[width=0.90\linewidth]{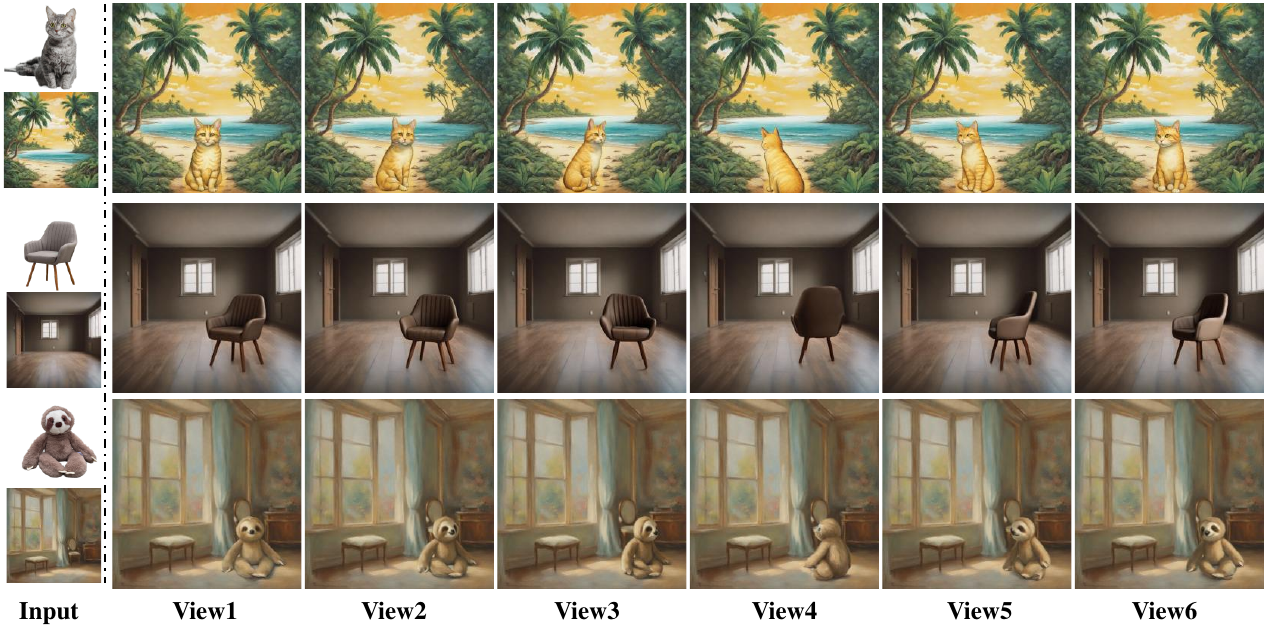} 
\caption{Visualization of view controllability. On the left are the input object images and backgrounds, and on the right are the generated results of \textit{FreeInsert}. Given an object image and a background, our \textit{FreeInsert} can seamlessly insert the object into the background with controllable view.}
\label{fig:visual_view}
\end{figure*}

\begin{figure*}[]
\centering
\includegraphics[width=0.90\linewidth]{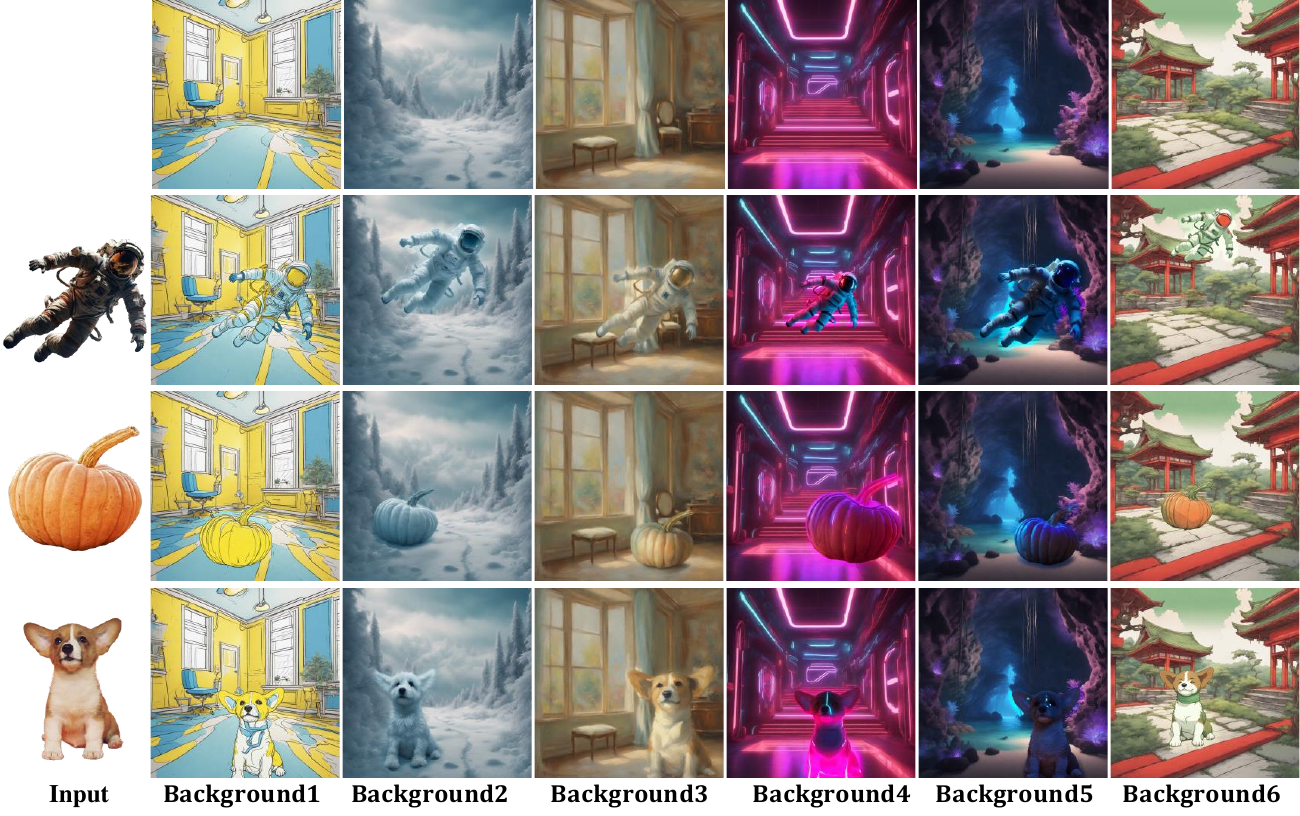} 
\caption{Visualization of style consistency. On the left are the input object images and on the right are the generated results of \textit{FreeInsert}. Given an object image and a background, our \textit{FreeInsert} can seamlessly insert the object into the background, maintaining style consistency.}
\label{fig:visual_style}
\end{figure*}

\section{Experiments}

\subsection{Benchmark}
Since there is no existing dataset to evaluate geometry-controllable customized image composition, we collected a dataset for evaluation, which includes object images and background images. For the object images, we collected 6 images from DreamBooth~\cite{ruiz2023dreambooth} and Image Sculpting~\cite{yenphraphai2024imagesculpture}. For images from DreamBooth~\cite{ruiz2023dreambooth}, we generated meshes using Image-to-3D generation model~\cite{liu2023threestudio}. For images from Image Sculpting~\cite{yenphraphai2024imagesculpture}, pre-generated meshes are provided. For each object, we rendered 6 images from different views. For some objects, we performed geometric edits on the meshes and rendered coarse images, resulting in a total of 40 rendered images. For the background images, we collected 20 images from MagicInsert~\cite{ruiz2024magicinsert}. Using the 40 rendered images as geometric conditions, we inserted the objects into the background images, ultimately generating 800 images for qualitative and quantitative evaluation. 

\subsection{Experimental Setup}
We use SDXL~\cite{podell2023sdxl} along with the corresponding depth-based ControlNet~\cite{zhang2023controlnet} and IP-Adapter~\cite{ye2023ipadapter} to implement our method. For the 3D generation phase, we use the Image-to-3D model provided by threestudio~\cite{liu2023threestudio}. Other high-quality 3D generation methods are also applicable to our framework. For the feature injection phase, we utilize all self-attention layers of the SDXL decoder and the first block of the SDXL upsampling decoder. Following Image Sculpting~\cite{yenphraphai2024imagesculpture}, we set $\tau_{q}=0.5T$, $\tau_{k}=0.5T$ and $\tau_{f}=0.2T$. We also use SDXL refiner after $t=0.1T$. For prompt generation, we use LLaVA-v1.5-7B~\cite{liu2024llava}. Our method is implemented in PyTorch and runs on a single A100 GPU.

\subsection{Comparisons}
We compared our method with TF-ICON~\cite{lu2023tficon} and AnyDoor~\cite{chen2024anydoor}. Since these two methods are solely designed for image composition and cannot control the geometry of the inserted objects, we used the original image $I_{obj}$ and the rendered image $I_{render}$ as object images for image composition, respectively.
\subsubsection{Qualitative Results}
We present the qualitative results in Figure \ref{fig:compare}. Our method successfully leverages the geometric conditions from 3D rendering to generate geometrically controlled object images that are stylistically consistent with the background images. TF-ICON~\cite{lu2023tficon} and AnyDoor~\cite{chen2024anydoor} produce relatively good composite images when the original object images are used as input. However, when rendered object images are used as input, the quality of their outputs deteriorates, and they sometimes fail to generate correct or reasonable images (column 4 and 6). This issue might be caused by artifacts in the rendered images. Therefore, it is difficult to achieve precise geometric control of objects in the generated images using these methods. As a comparison, our method is not constrained by the artifacts in rendered images and can generate high-quality images. Additionally, TF-ICON tends to alter the background image during composition, whereas our method maintains the background image more effectively. As shown in Figure \ref{fig:compare}, TF-ICON causes the hue of the image background to change, which is not what we expect. AnyDoor tends to preserve the original style of the object, lacking the ability to ensure style consistency with the background image. 

\begin{figure}[t]
\centering
\includegraphics[width=0.99\linewidth]{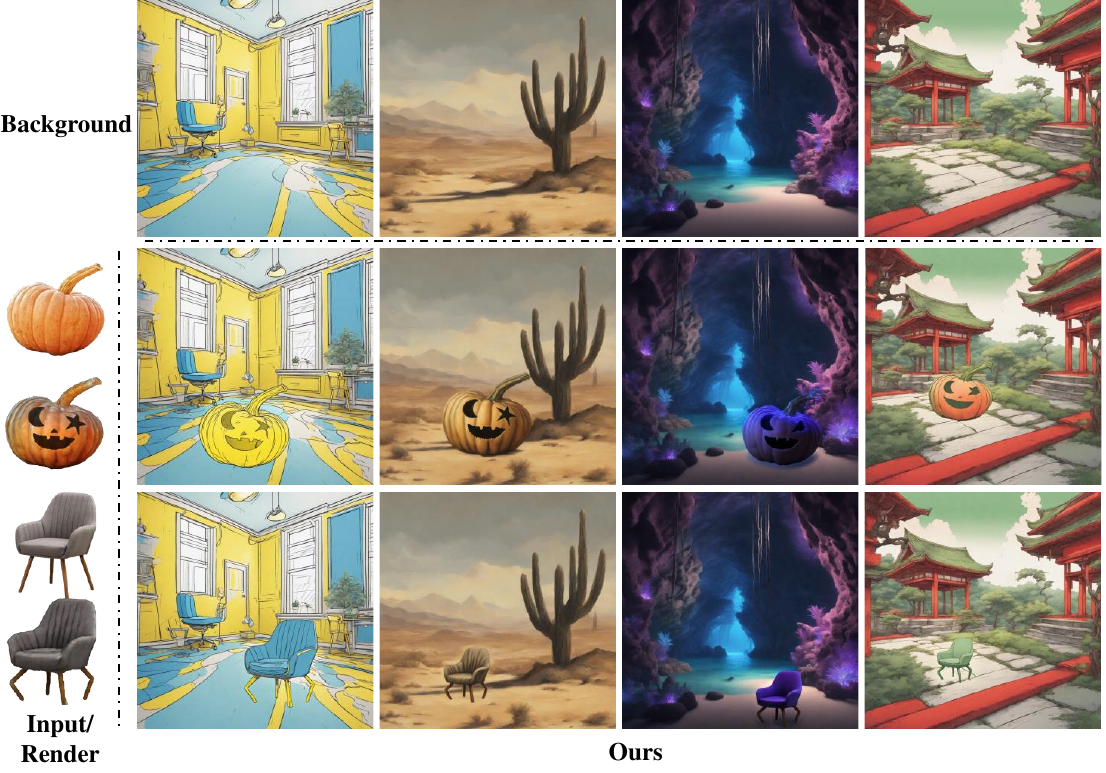} 
\vspace{-3mm}
\caption{Visualization of geometric controllability and style consistency. On the left are the input object images and rendered images, and on the right are the generated results of \textit{FreeInsert}. Given an object image and a background, our \textit{FreeInsert} can seamlessly insert the object into the background and can control the object's geometry and style.}
\label{fig:visual_geo_style}
\end{figure}

We also evaluated the geometric and style control capability of our proposed method. As shown in Figure \ref{fig:visual_view}, given a background image and an object, our method can successfully achieve view controllable object insertion into arbitrary scenes by adjusting the rendering view.  As shown in Figure \ref{fig:visual_style}, our method also show great performance about style consistency. Given the same background image, our method is unaffected by the original style of the input objects and can synthesize new images with a style consistent with the background, while ensuring that geometric control remains consistent with the rendered image. As shown in Figure \ref{fig:visual_geo_style}, our method can also control the object’s geometry and style. Users can can customize the shape of objects according to their needs.

\subsubsection{Quantitative Results }

In the quantitative evaluation, we calculate the CLIP similarity, DINO similarity and LPIPS between the generated objects in the target region and the input objects to assess the object fidelity of the synthesized content ($\text{CLIP}_{obj}$, $\text{DINO}_{obj}$ and $\text{LPIPS}_{obj}$). We compute the CLIP similarity, DINO similarity and LPIPS between the target region and the background image to evaluate the style consistency of the synthesized image ($\text{CLIP}_{style}$, $\text{DINO}_{style}$ and $\text{LPIPS}_{style}$). 

The experimental results are shown in Table \ref{tab:compare}. Our method achieves better $\text{CLIP}_{style}$, $\text{DINO}_{style}$ and $\text{LPIPS}_{style}$ scores compared to other methods, demonstrating its advantage in style consistency. The $\text{CLIP}_{obj}$ and $\text{DINO}_{obj}$ scores are slightly lower than those of other methods, which may be due to style information leakage. $\text{LPIPS}_{obj}$ is comparable to other methods. Overall, our method achieves a better balance between object fidelity and style consistency.

\subsection{Ablation Study}
\begin{figure*}[t]
\centering
\includegraphics[width=0.90\textwidth]{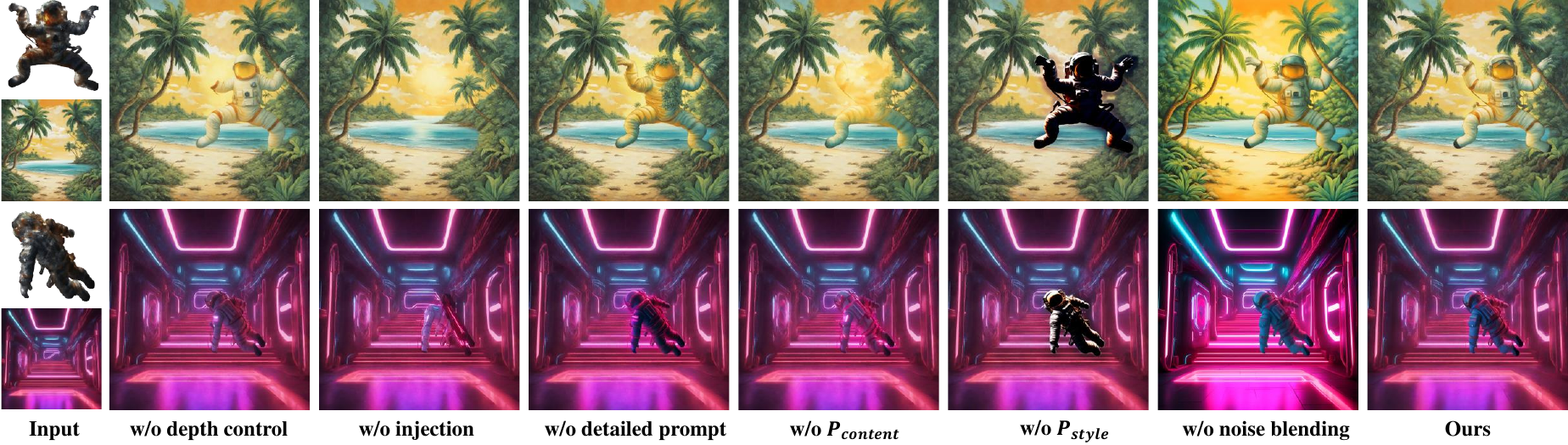} 
\vspace{-3mm}
\caption{Visual comparisons on the effectiveness of geometry, content and style control. On the left are rendered images and backgrounds and on the right are the generated results of different ablation studies (Geometry control: column 2, 3, Content control: column 4, 5, Style control: column 6, 7).}
\label{fig:ablation}
\end{figure*}

We have conducted experiments to validate the effectiveness of the proposed geometry, content, and style control. For geometry control, we test the efficacy of depth control and feature injection. Regarding content control, we replace the detailed prompt with simple text such as ``a photo of a $\langle \cdot \rangle$”, and also evaluate the effectiveness of $P_{content}$. For style control, we evaluate the effectiveness of $P_{style}$ and noise blending strategy. In our ablation study, we select the object ``astronaut” for all experiments. We render 6 images from different views and 2 images with different poses, resulting in 8 images for geometric control. The ``astronaut" object is inserted into 20 different background images with 8 geometric control and generate 160 images in total. We use $\text{CLIP}_{obj}$ and $\text{CLIP}_{style}$ to evaluate content consistency with $I_{obj}$ and style consistency with $I_{bg}$, respectively. To measure the effectiveness of geometric control, we calculated D-RMSE for quantitative assessment. D-RMSE is calculated by the root mean square error between the depth maps of the rendered images and generated images using MiDaS~\cite{xu2021midas}. D-RMSE is the lower the better. The experimental results are shown in Figure \ref{fig:ablation} and Table \ref{tab:ablation}.

\noindent \textbf{Geometry Control.}
For geometry control, both removing the depth control and feature injection components result in an increased D-MRSE, thereby compromising the accuracy of geometric control. This effect is particularly pronounced when the feature injection module is removed, significantly reducing geometric control accuracy. As illustrated in Figure \ref{fig:ablation}, it becomes nearly impossible to generate discernible objects (column 3). The removal of depth control leads to the generation of incomplete objects, such as the astronaut's arm in column 2.

\noindent \textbf{Content Control.}
Regarding the content control, replacing detailed prompt with simple text and removing $P_{content}$ both lead to a reduction in $\text{CLIP}_{obj}$, thereby diminishing the fidelity to the original inserted objects. As shown in Figure \ref{fig:ablation}, using simple prompt instead of detailed prompt results in a noticeable loss of detail in the generated images. Since the initial features in the diffusion process primarily originate from the injection module, removing the content embedding impedes the effective transmission of content features to the diffusion process, resulting in generated content that is not faithful to the original objects.

\begin{table*}[t]
\vspace{-3mm}
\caption{Quantitative comparisons with existing methods.}
\label{tab:compare}
\centering
\vspace{-3mm}
\begin{tabular}{lcccccc}
\toprule
Methods & $\text{CLIP}_{obj} \uparrow$ & $\text{DINO}_{obj} \uparrow$ & $\text{CLIP}_{style} \uparrow$ & $\text{DINO}_{style} \uparrow$ & $\text{LPIPS}_{obj} \downarrow$ & $\text{LPIPS}_{style} \downarrow$  \\
\midrule
${\text{TF-ICON}}_{object}$ & 0.822 & 0.598 & 0.600 & 0.158 & 0.652 & 0.253\\
${\text{TF-ICON}}_{render}$ & 0.803 & 0.571 & 0.615 & 0.184 & 0.651 & 0.241 \\
${\text{AnyDoor}}_{object}$ & 0.886 & 0.649 & 0.608 & 0.170 & 0.645 & 0.221 \\
${\text{AnyDoor}}_{render}$ & 0.822 & 0.579 & 0.610 & 0.175 & 0.642 & 0.221 \\
Ours & 0.784 & 0.517 & 0.699 & 0.317 & 0.650 & 0.120 \\
\bottomrule
\end{tabular}
\end{table*}

\begin{table}[t]
\caption{Ablation studies on the geometry, content and style control we proposed.}
\label{tab:ablation}
\centering
\vspace{-3mm}
\begin{tabular}{clccc}
\toprule
\multicolumn{2}{c}{Methods} & $\text{CLIP}_{obj}$ & $\text{CLIP}_{style}$ & $\text{D-RMSE}$ \\
\midrule
\multirow{2}{*}{Geometry} & w/o depth control & 0.873 & 0.729 & 0.254 \\
 & w/o injection & 0.820 & 0.754 & 0.351 \\
\multirow{2}{*}{Content} & w/o detailed prompt & 0.874 & 0.726 & 0.215 \\
 & w/o $P_{content}$ & 0.845 & 0.732 & 0.257 \\
\multirow{2}{*}{Style} & w/o $P_{style}$ & 0.899 & 0.666 & 0.204 \\
 & w/o noise blending & 0.892 & 0.725 & 0.222 \\
 & Ours & 0.891 & 0.720 & 0.214 \\
\bottomrule
\end{tabular}
\end{table}

\begin{figure}[t]
\centering
\vspace{-3mm}
\includegraphics[width=0.86\linewidth]{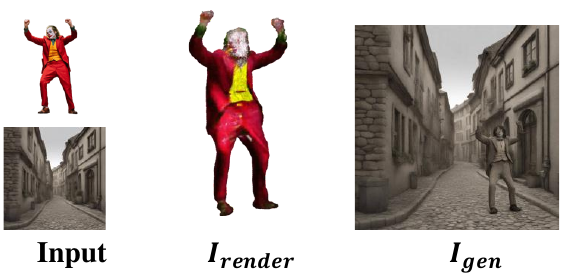} 
\vspace{-3mm}
\caption{Failure case.}
\label{fig:failure}
\end{figure}

\noindent \textbf{Style Control.}
In terms of style control, the removal of $P_{style}$ significantly reduces $\text{CLIP}_{style}$. As depicted in Figure \ref{fig:ablation}, the generated images lack stylistic consistency with the background images. Although not using noise blending does not decrease $\text{CLIP}_{style}$, it causes a change in the content and color of background image, which is undesirable, as shown in Figure \ref{fig:ablation}. Without noise blending, the grass on the ground (row 1) and the ceiling (row 2) have changed.

\subsection{Limitations}
The proposed method struggles with high-fidelity object insertion, such as humans. As shown in Figure \ref{fig:failure}, while we aim to preserve fine-grained details (e.g., the joker) in \( I_{gen} \), the rendered image's poor quality fails to capture these details. This is due to low-quality 3D assets generated by the image-to-3D model, especially in critical regions like the face. Additionally, since the content information is injected through cross-attention mechanisms, which do not support fine-grained control, the facial details are prone to undesired alterations. In future work, we plan to explore more advanced and detail-preserving techniques to address the challenges associated with high-fidelity object insertion.

\section{Conclusion}
In this paper, we introduce the task of personalized object insertion with geometry and style control and propose \textit{FreeInsert} as a solution. Our approach is a training-free method based on diffusion. In our approach, the object is first converted from 2D to 3D for precise geometric editing and view choosing, and then projected back to 2D as geometric control. Our pipeline integrates geometry, content, and style controls to produce geometrically controlled and style-consistent composite images. Specifically, we utilize depth controlnet and feature injection for geometric control, fine-grained text prompts combined with a content adapter for content control, and an adapter with style embedding for style control. Additionally, to ensure the background image unchanged, we employ a noise blending strategy.

\begin{acks}
This work was partly supported by the NSFC62431015, the Fundamental Research Funds for the Central Universities, Shanghai Key Laboratory of Digital Media Processing and Transmission under Grant 22DZ2229005, 111 project BP0719010.
\end{acks}

\bibliographystyle{ACM-Reference-Format}
\balance
\bibliography{reference}

\end{document}